\pdfoutput=1

\documentclass[11pt]{article}

\usepackage[]{emnlp2021}

\usepackage{times}
\usepackage{latexsym}

\usepackage{booktabs}
\usepackage{tabularx}
\usepackage{lipsum}
\usepackage{graphicx}
\usepackage{todonotes}
\usepackage{svg}
\usepackage{amsmath}
\usepackage{multirow}

\usepackage[T1]{fontenc}

\usepackage[utf8]{inputenc}

\usepackage{microtype}

\title{Label Verbalization and Entailment\\for Effective Zero- and Few-Shot Relation Extraction}

\author{Oscar Sainz \hspace{15pt} Oier Lopez de Lacalle \hspace{15pt} Gorka Labaka \\ {\bf Ander Barrena} \hspace{15pt} {\bf Eneko Agirre} \\ \\
HiTZ Basque Center for Language Technologies - Ixa NLP Group\\
University of the Basque Country (UPV/EHU)\\
\{oscar.sainz, oier.lopezdelacalle, gorka.labaka, ander.barrena, e.agirre\}@ehu.eus
}

\begin{document}
\maketitle
\begin{abstract}
Relation extraction systems require large amounts of labeled examples which are costly to annotate.
In this work we reformulate relation extraction as an entailment task, with simple, hand-made, verbalizations of relations produced in less than 15 minutes per relation. 
The system relies on a pretrained textual entailment engine which is run as-is (no training examples, zero-shot) or further fine-tuned on labeled examples (few-shot or fully trained). 
In our experiments on TACRED we attain 63\% F1 zero-shot, 69\% with 16 examples per relation (17\% points better than the best supervised system on the same conditions), and only 4 points short of the state-of-the-art (which uses 20 times more training data).
We also show that the performance can be improved significantly with larger entailment models, up to 12 points in zero-shot, giving the best results to date on TACRED when fully trained. 
The analysis shows that our few-shot systems are especially effective when discriminating between relations, and that the performance difference in low data regimes comes mainly from identifying no-relation cases.
\end{abstract}

\section{Introduction}



Given a context where two entities appear, the Relation Extraction (RE) task aims to predict the semantic relation (if any) holding between the two entities. Methods that fine-tune large pretrained language models (LM) with large amounts of labelled data have  established the state of the art ~\cite{yamada-etal-2020-luke}. Nevertheless, due to differing languages, domains and the cost of human annotation,  there is typically a very small number of labelled examples in real-world  applications, and such models perform poorly \cite{schick-schutze-2021-exploiting}. 

As an alternative, methods that only need a few examples (few-shot) or no examples (zero-shot) have emerged. 
For instance, \textit{prompt based learning} proposes hand-made or automatically learned task and label verbalizations \cite{puri2019zeroShot,schick-schutze-2021-exploiting,schick2020small} as an alternative to standard fine-tuning \cite{gao2020making,scao2021data}. In these methods, the prompts are input to the LM together with the example, and the language modelling objective is used in learning and inference. 
In a different direction, some authors reformulate the target task (e.g. document classification) as a  \textit{pivot task} (typically question answering or textual entailment), which allows the use of readily available question answering (or entailment) training data \cite{yin-etal-2019-benchmarking,levy-etal-2017-zero}.  In all cases, the underlying idea is to cast the target task into a formulation which allows us to exploit the knowledge implicit in pre-trained LM (prompt-based) or general-purpose question answering or entailment engines (pivot tasks). 
Prompt-based approaches are very effective when the label verbalization is given by one or two words (e.g. text classification), as they can be easily predicted by language models, but strive in cases where the label requires a more elaborate description, as in RE. We thus \textbf{propose to reformulate RE as an entailment problem}, where the verbalizations of the relation label are used to produce a hypothesis to be confirmed by an off-the-shelf entailment engine.

In our work\footnote{Code and splits available at: \url{https://github.com/osainz59/Ask2Transformers}} we have manually constructed verbalization templates for a given set of relations. Given that some verbalizations might be ambiguous (between city of birth and country of birth, for instance) we complemented them with entity type constraints. In order to ensure that the manual work involved is limited and practical in real-world applications, we allowed at most 15 minutes of manual labor per relation. The verbalizations are used as-is for zero-shot RE, but we also recast labelled RE examples as entailment pairs and fine-tune the entailment engine for few-shot RE.


The results on the widely used TACRED~\cite{zhang2017tacred} RE dataset in zero- and few-shot scenarios are excellent, well over state-of-the-art systems using the same amount of data. In addition our method scales well with large pre-trained LMs and large amounts of training data, reporting the best results on TACRED to date.

\section{Related Work}
\label{sec:related}

\paragraph{Textual Entailment.} It was first presented by~\citet{10.1007/11736790_9} and further developed by ~\citet{bowman-etal-2015-large} who called it Natural Language Inference (NLI). Given a textual premise and hypothesis, the task is to decide whether the premise entails or contradicts (or is neutral to) the  hypothesis. The current state-of-the-art uses large pre-trained LM fine-tuned in NLI datasets~ \cite{Lan2020ALBERT:,roberta, conneau-etal-2020-unsupervised, lewis-etal-2020-bart, deberta}.


\paragraph{Relation Extraction.} The best results to date on RE are obtained by fine-tuning large pre-trained language models equipped with a classification head. ~\citet{joshi-etal-2020-spanbert} pretrains a masked language model on random contiguous spans to learn span-boundaries and predict the entire masked span. LUKE \cite{yamada-etal-2020-luke} further pretrains a LM predicting entities from Wikipedia, and  using entity information as an additional input embedding layer. K-Adapter~\cite{wang2020kadapter} fixes the parameters of the pretrained LM and use Adapters to infuse factual and linguistic knowledge from Wikipedia and dependency parsing.



TACRED~\cite{zhang2017tacred} is the largest 
and most widely used dataset for RE in English. It is derived from the TAC-KBP relation set, with labels  obtained via crowdsourcing. Although alternate versions of TACRED have been published recently~\cite{alt-etal-2020-tacred, retacred}, the state of the art is mainly tested in the original version. 



\paragraph{Zero-Shot and Few-Shot learning.} ~\citet{brown2020language} showed that task descriptions (\emph{prompts}) can be fed into LMs for task-agnostic and few-shot performance. In addition, \cite{schick2020small, schick-schutze-2021-exploiting, tam2021AdaPET} extend the method and allow finetuning of LMs on a variety of tasks. 
Prompt-based prediction treats the downstream task as a (masked) language modeling problem, where the model directly generates a textual response to a given  prompt. 
The manual generation of effective prompts is costly and requires domain expertise. \citet{gao2020making} provide an effective way to generate prompts for text classification tasks that surpasses the performance of hand picked ones. The approach uses few-shot training with a generative T5 model~\cite{JMLR:v21:20-074} to learn to decode effective prompts. 
Similarly,  \citet{liu2021gpt} automatically search prompts in a embedding space which can be simultaneously fine-tuned along with the pre-trained language model. Note that previous prompt-based models run their zero-shot models on a semi-supervised setting in which some amount of labeled data is given in training. 
Prompts can be easily generated for text classification. Other tasks require more elaborate templates ~\cite{goswami-etal-2020-unsupervised,li2021documentlevel} and currently no effective prompt-based methods for RE exist.


Besides prompt-based methods, the use of pivot tasks has been widely use for few/zero-shot learning. For instance, relation and event extraction have been cast as a question answering problem ~\cite{levy-etal-2017-zero,du-cardie-2020-event}, associating each slot label to at least one natural language question. Closer to our work, NLI has been shown too to be a successful pivoting task for text classification~\cite{yin-etal-2019-benchmarking, yin-etal-2020-universal, facebook_entailment, sainz-rigau-2021-ask2transformers}. These works verbalize the labels, and apply an entailment engine to check whether the input text entails the label description. 

In similar work to ours, the relation between entailment and RE was explored by  \citet{obamuyide-vlachos-2018-zero}. In their work they present some preliminary experiments where they cast RE as entailment, but only evaluate performance as binary entailment, not as a RE task. As a consequence they do not have competing positive labels and avoid RE inference and the issue of detecting no-relation.




\paragraph{Partially vs. fullly unseen labels in RE.} 
Existing zero/few-shot RE models usually see some labels during training (\emph{label partially unseen}), which helps generalize to the unseen label~\cite{levy-etal-2017-zero, obamuyide-vlachos-2018-zero, han-etal-2018-fewrel, chen2021zsbert}. These approaches do not fully address the data scarcity problem. In this work we address the more challenging \emph{label fully unseen} scenario.

    

    

    

\begin{figure*}[ht]
    \centering
    \resizebox{\textwidth}{!}{
        \includegraphics{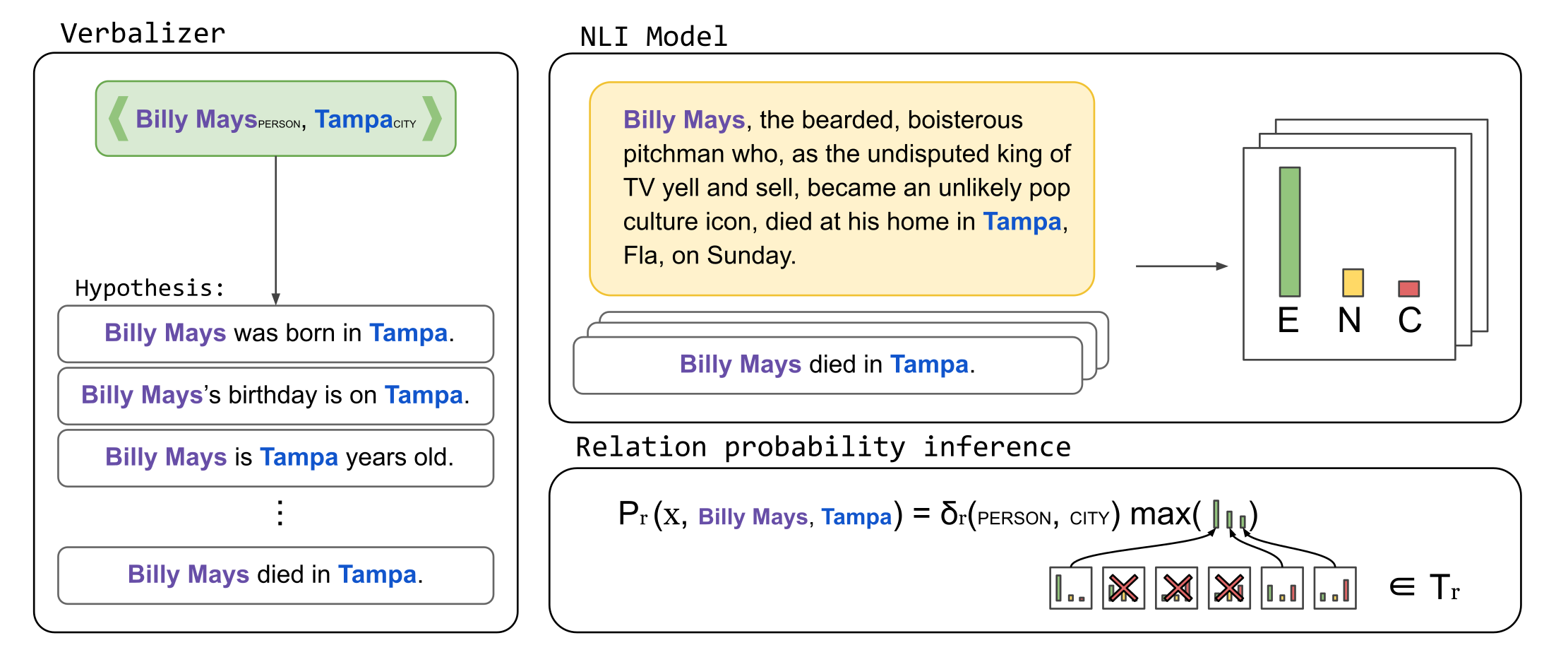}
    }
    \caption{General workflow of our entailment-based RE approach.}
    \label{fig:architecture}
\end{figure*}

\section{Entailment for RE}
\label{sec:systems}

In this section we describe our models for zero- and few-shot RE. 



\subsection{Zero-shot relation extraction} 
\label{ssed:zeroshot}

We reformulate RE as an entailment task: given the input text containing the two entity mentions as the premise and the verbalized description of a relation as hypothesis, the task is to infer if the premise entails the hypothesis according to the NLI model. 
Figure \ref{fig:architecture} illustrates the main 3 steps of our system. The first step is focused on relation verbalization to generate the set of hypotheses. In the second we run the NLI model\footnote{We describe the NLI models in Section \ref{ssec:nli_models}} and obtain the entailment probability for each hypothesis. 
 Finally, based on the probabilities and the entity types, we return the relation label that maximizes the probability of the hypothesis, including the \textsc{no-relation} label.


\paragraph{Verbalizing relations as hypothesis.}
The hypotheses are automatically generated using a set of templates. Each template verbalizes the relation holding between two entity mentions. 
For instance, the relation \textsc{per:date\_of\_birth} can be verbalized with the following template: \texttt{\{subj\}'s birthday is on \{obj\}}. More formally, given the text $x$ that contains the mention of two entities ($x_{e1}$, $x_{e2}$) and template $t$, the hypothesis $h$ is generated by $\textsc{verbalize}(t,x_{e1},x_{e2})$, which substitutes the \texttt{subj} and \texttt{obj} in the $t$ with the entities $x_{e1}$ and $x_{e2}$, respectively\footnote{Note that the entities are given in a fixed order, that is the relation needs to hold between  $x_{e1}$ and $x_{e2}$ in that order; the reverse ($x_{e2}$ and $x_{e1}$) would be a different example.}. 
Figure \ref{fig:architecture} 
shows four verbalizations for the given entity pair. 

A relation label can be verbalized by one or more templates. For instance, in addition to the previous template, \textsc{per:date\_of\_birth} is also verbalized with \texttt{\{subj\} was born on \{obj\}}. 
At the same time, a template can verbalize more than one relation label. For example, \texttt{\{subj\} was born in \{obj\}} verbalizes \textsc{per:country\_of\_birth} and \textsc{per:city\_of\_birth}. 
In order to cope with such ambiguous verbalizations, we added the entity type information to each relation, e.g. \texttt{COUNTRY} and \texttt{CITY}  for each of the relations in the previous example. 
\footnote{Alternatively, one could think on more specific verbalizations, such as  \texttt{\{subj\} was born in the city of \{obj\}} for \textsc{per:city\_of\_birth}. In the checks done in the available 15 min. such specific verbalizations had very low recall and were not finally selected.} 


We defined a function $\delta_r$ for every relation $r \in R$ that checks the entity coherence between the template and the current relation label: 
\[   
   \delta_r(e_1, e_2) = 
        \begin{cases} 
        1 & e_1 \in E_{r1} \land e_2 \in E_{r2} \\
        0 & \text{otherwise} 
    \end{cases}
\]

\noindent where $e_1$ and $e_2$ are the entity types of the first and second arguments, $E_{r1}$ and $E_{r2}$ are the set of allowed types for the first and second entities in relation $r$. This function is used at inference time, to discard relations that do not match the given types.
Appendix \ref{sec:templates} lists all templates and entity type restrictions used in this work. 

\paragraph{NLI for inferring relations.}
In a second step we make use of the NLI model to infer the relation label. Given the text $x$ containing two entities $x_{e1}$ and $x_{e2}$ the system returns the relation $\Hat{r}$ from the set of possible relation labels $R$ with the highest entailment probability as follows:

\begin{equation}
\label{eq1}
    \Hat{r} = \arg \max_{r \in R} \mathsf{P}_r(x, x_{e1}, x_{e2}) 
\end{equation}

The probability of each relation $\mathsf{P}_r$ 
is computed as the probability of the hypothesis 
that yields the maximum entailment probability (Eq.~\ref{eq2}), among the set of possible hypothesis. In case the two entities do not match the required entity types, the probability would be zero.

\begin{multline}
\label{eq2}
    \mathsf{P}_r(x, x_{e1}, x_{e2}) = 
    \delta_r(e_1, e_2) \max_{t \in T_r} \mathsf{P}_{NLI}(x,\textit{hyp}) \\ 
    \textit{where} \,\textit{hyp} = \textsc{verbalize}(t,x_{e1}, x_{e2}) 
\end{multline}

\noindent where $\mathsf{P}_{NLI}$ is the entailment probability between the input text and the hypothesis generated by the template verbalizer. Although entailment models return probabilities for entailment, contradiction and neutral, $\mathsf{P}_{NLI}$ just makes use of the entailment probability\footnote{The probabilities for relations $\mathsf{P}_r$ defined in Eq.~\ref{eq2} are independent from each other, which, in a way, they could be easily extended to multi-label  classification task.}.
The right hand-side of Figure \ref{fig:architecture} 
shows the application of NLI models and how the probability for each relation,  $\mathsf{P}_r$, is computed. 



\paragraph{Detection of no-relation.} In supervised RE, the \textsc{no-relation} case is taken as an additional label. In our case we examined two approaches.

In \textbf{template-based detection} we propose an additional template
as if it was yet another relation label, and treated it as another positive relation in Eq.~\ref{eq1}. The template for \textsc{no-relation}: \texttt{\{subj\} and \{obj\} are not related}.

In \textbf{threshold-based detection} we apply a threshold $\mathcal{T}$ to $\mathsf{P}_r$ in Eq. \ref{eq2}.  
If none of the relations surpasses the threshold, then our system returns \textsc{no-relation}. On the contrary, the model returns the relation label of highest probability (Eq.~\ref{eq1}).
When no development data is available, the threshold $\mathcal{T}$ is set to 0.5. 
Alternatively, we estimate $\mathcal{T}$ using the available development dataset, as described in the experimental part.


\subsection{Few-Shot relation extraction}

Our system is based on a NLI model which has been pretrained on annotated entailment pairs. When labeled relation examples exist, we can reformulate them as labelled NLI pairs, and use them to fine-tune the NLI model to the task at hand, that is, assigning highest entailment probability to the verbalizations of the correct relation, and assigning low entailment probabilities to the rest of the hypothesis (see Eq. \ref{eq2}).



Given a set of labelled relation examples, we use the following steps to produce labelled entailment pairs for fine-tuning the NLI model. 1) For each \textbf{positive} relation example we generate at least one \textbf{entailment} instance with the templates that describes the current relation. That is, we generate one or several premise-hypothesis pairs labelled as entailment. 2) For each \textbf{positive} relation example we generate one \textbf{neutral} premise-hypothesis instance, taken at random from the templates that do not represent the current relation. 3) For each \textbf{negative} relation example we generate one \textbf{contradiction} example, taken at random from the templates of the rest of relations.


If a template is used for the no-relation case, we do the following: 
 First, for each \textbf{no-relation} example we generate one \textbf{entailment} example with the no-relation template. Then, for each \textbf{positive} relation example we generate one \textbf{contradiction} example using the no-relation template.




\section{Experimental Setup}

\begin{table*}
    \centering
    \resizebox{0.8\textwidth}{!}{
        \begin{tabular}{cc|rrrrrrrrr}
            \toprule
             &  & \multicolumn{3}{c}{Train (Gold)} & \multicolumn{3}{c}{Train (Silver)} & \multicolumn{3}{c}{Development} \\
             & & \multicolumn{2}{c}{\# Pos} & \# Neg & \multicolumn{2}{c}{\# Pos} & \# Neg & \multicolumn{2}{c}{\# Pos} & \# Neg \\
             Scenario & Split & mean & total & total  & mean & total & total  & mean & total & total \\
            \midrule
            \midrule
            Full training & 100\% & 317.4 & 13013 & 55112 & - & - & - & 132.6 & 5436 & 17195 \\
            \midrule
            \multirow{2}{*}{Zero-Shot} & No Dev & - & - & - & - & - & - & 0 & 0 & 0 \\
                                       & 1\% Dev& - & - & - & - & - & - & 1.9 & 54 & 173 \\
            \midrule
            \multirow{3}{*}{Few-Shot} & 1\% & 3.6 & 130 & 552 & - & - & - & 1.9 & 54 & 173 \\
                                      & 5\% & 16.3 & 651 & 2756 & - & - & - & 7.0 & 272 & 861 \\
                                      & 10\% & 32.6 & 1302 & 5513 & - & - & - & 13.6 & 544 & 1721 \\
            \midrule
            \multirow{4}{*}{Data Augment.} & 0\% & 0 & 0 & 0 & 246.3  & 9850 & 41205 & 1.9 & 54 & 173 \\
             & 1\% & 3.6 & 130 & 552 & 246.3  & 9850 & 41205 & 1.9 & 54 & 173 \\
                                      & 5\% & 16.3 & 651 & 2756 & 246.3  & 9850 & 41205 & 7.0 & 272 & 861 \\
                                      & 10\% & 32.6 & 1302 & 5513 & 246.3  & 9850 & 41205 & 13.6  & 544 & 1721 \\
            \bottomrule
        \end{tabular}
    }
    \caption{Statistics about the dataset scenarios based on TACRED used in the paper, including positive examples per relation, total amount of positive examples and the total amount of negative (no-relation) examples.}
    \label{tab:scenarios}
\end{table*}


In this section we describe the dataset and scenarios we have used for evaluation, how we performed the verbalization process, the different pre-trained NLI models we have used and the state-of-the-art baselines that we compare with.

\subsection{Dataset and scenarios}

We designed three different low-resource scenarios based on the large-scale TACRED \cite{zhang2017tacred} dataset. The full dataset consists of 42 relation labels, including the \textsc{no-relation} label, and each example contains the information about the entity type, among other linguistic information. 
The scenarios are described in Table \ref{tab:scenarios} and are formed by different splits of the original dataset. We applied a stratified sampling method to keep the original label distribution. 



\paragraph{Zero-Shot.} The aim of this scenario is the evaluation of the models when no data is available for training. We present two different  situations on this scenario: 1) no data is available for 
development (0\% split) and 2) a small development set is available with around 2 examples per relation (1\% split)\footnote{This setting is comparable to one where the examples in the guidelines are used as development.}. In this scenario the models are not allowed to train their own parameters but development data is used to adjust the hyperparameters.

\begin{table*}[!ht]
    \centering
    \resizebox{.825\textwidth}{!}{
        \begin{tabular}{l|rc|ccc|ccc} 
            \toprule
             & & MNLI & \multicolumn{3}{c|}{No Dev ($\mathcal{T}=0.5)$} & \multicolumn{3}{c}{1\% Dev} \\
            NLI Model & \# Param. & Acc. & Pr. & Rec. & F1 & Pr. & Rec. & F1 \\ 
            \midrule
            \midrule
            ALBERT\textsubscript{xxLarge} & 223M & 90.8
            & 32.6 & \textbf{79.5} & 46.2 & 55.2 & 58.1 & 56.6 \small{$\pm 1.4$} \\ 
            RoBERTa & 355M & 90.2
            & 32.8 & 75.5 & 45.7 & 58.5 & 53.1 & 55.6 \small{$\pm 1.3$} \\ 
            BART & 406M & 89.9
            & 39.0 & 63.1 & 48.2  & 60.7 & 46.0 & 52.3 \small{$ \pm 1.8$} \\ 
            DeBERTa\textsubscript{xLarge} & 900M & 91.7
            & 40.3 & 77.7 & 53.0 & \textbf{66.3} & 59.7 & \textbf{62.8} \small{$\pm 1.7$} \\ 
            DeBERTa\textsubscript{xxLarge} & 1.5B & 91.7
            & \textbf{46.6} & 76.1 & \textbf{57.8} & 63.2 & \textbf{59.8} & 61.4 \small{$\pm 1.0$} \\ 
            \bottomrule
        \end{tabular}
    }
    \caption{Zero-Shot scenario results (Precision, Recall and F1) for our system using several pre-trained NLI models in two settings: no development  (default threshold $\mathcal{T}$=0.5), and small development (1\% Dev.) for setting $\mathcal{T}$. In the leftmost columns we report the number of parameters and the accuracy in MNLI. For the 1\% setting we report the median measures along with the F1 standard deviation in 100 runs. }
    \label{tab:zero_shot}
\end{table*}

\paragraph{Few-Shot.} 
This scenario presents the challenge of solving the RE task with just a few examples per relation. 
We present three settings commonly used in few-shot learning~\cite{gao2020making} \footnote{The commonly reported value in few-shot scenarios is 16 examples per label. We also added the 3-8 and 32 examples settings in the evaluation.}: around 4 examples per relation (1\% of the training data in TACRED), around 16 examples per relation (5\%) and around 32 examples per relation (10\%). We reduced the development set following the same ratio.

\paragraph{Full Training.} In this setting we use all available training and development data.

\paragraph{Data Augmentation.} 
In this scenario we want to test whether a silver dataset produced by running our systems on untagged data can be used to train a supervised relation extraction system (cf. Section \ref{sec:systems}). In this scenario 75\% of the training data in TACRED is set aside as unlabeled data\footnote{We use part of the original TACRED dataset to produce silver data in order not to introduce noise coming from different documents and/or pre-processing steps.}, and the rest of the training data is used in different splits (ranging from 1\% to 10\%). 
Under this setting we carried out two type of experiments: In the zero-shot experiments (0\% in the table) we use our NLI based model
to annotate the silver data and then fine-tune the RE model exclusively on the silver data. In the few-shot experiments the NLI model is first fine-tuned with the gold data, then used to annotate the silver data and finally the RE model is fine-tuned over both, silver and gold, annotations.

\subsection{Hand-crafted relation templates} 
\label{ssec:templates}

We manually created the templates to verbalize relation labels, based on the TAC-KBP guidelines which underlie the TACRED dataset. We limited the time for creating the templates of each relation to less than 15 minutes. Overall, we created 1-8 templates per relation (2 on average) (cf. Appendix \ref{sec:templates} for full list).



The verbalization process consists of generating one or more templates that describe the relation and contain the placeholders \texttt{\{subj\}} and \texttt{\{obj\}}. The developer building the templates was given the task guidelines (brief description of the relation, including one or two examples and the type of the entities) and a NLI model (\textit{roberta-large-mnli} checkpoint). For a given relation, he/she would create a template (or set of templates) and check whether the NLI model is able to output a high entailment probability for the template when applied on the guideline example(s). He/she could run this process for any new template that he/she could come up with. There was no strict threshold involved for selecting the templates, just the intuition of the developer. The spirit was to come up with simple templates quickly, and not to build numerous complex templates or to optimize entailment probabilities.



\subsection{Pre-Trained NLI models}
\label{ssec:nli_models}

For our experiments we tried different NLI models that are publicly available with the Hugging Face  Transformers \cite{wolf-etal-2020-transformers} python library. We tested the following models which implement different architectures, sizes and pre-training objectives and were fine-tuned mainly over the MNLI \cite{williams-etal-2018-broad} dataset\footnote{ALBERT was trained in some additional NLI datasets.}: ALBERT \cite{Lan2020ALBERT:}, RoBERTa \cite{roberta}, 
BART \cite{lewis-etal-2020-bart} and DeBERTa v2 \cite{deberta}. Table \ref{tab:zero_shot} reports the number of parameters of these models. Further details on models can be found in Appendix~\ref{sec:pretrained}.

For each of the scenarios we have tested different models. In zero-shot and full training scenarios we compare all the pre-trained models using the templates described in Section \ref{ssec:templates}. For few-shot we used RoBERTa for comparability, as it was used in 
state-of-the-art systems (cf. Section \ref{ssec:RE_models}), and DeBERTa which is the largest NLI model available on the HUB\footnote{\url{https://huggingface.co/models}}. Finally, we only tested RoBERTa in data-augmentation experiments.

We ran 3 different runs on each of the experiments using different random seeds. In order to make a fair comparison with state-of-the-art systems (cf section 4.4.), we performed a hyperparameter exploration in the full training scenario, using the resulting configuration also in the zero/few-shot scenarios.
We fixed the batch size at 32 for both RoBERTa and DeBERTa, and search the optimum learning-rate among $\{1e^{-6}, 4e^{-6}, 1e^{-5}\}$ on the development set. The best results were obtained using $4e^{-6}$ as learning-rate. For more detailed information refer to the Appendix \ref{sec:experimental_details}.


\subsection{State-of-the-art RE models}
\label{ssec:RE_models}

We compared the NLI approach with the systems reporting the best results to date on TACRED:  SpanBERT \cite{joshi-etal-2020-spanbert},  K-Adapter \cite{wang2020kadapter} and LUKE \cite{yamada-etal-2020-luke} (cf. Section~\ref{sec:related}). In addition, we also report the results obtained by the vanilla RoBERTa baseline proposed by \citet{wang2020kadapter} that serves as a reference for the improvements.  We re-trained the different systems on each scenario setting using their publicly available implementations and best performing hyperparameters reported by the authors. All these models have a comparable number of parameters.


\begin{table*}[t]
    \centering
    \resizebox{0.85\textwidth}{!}{
        \begin{tabular}{l|ccc|ccc|ccc}
            \toprule
             & \multicolumn{3}{c}{1\%} & \multicolumn{3}{c}{5\%} & \multicolumn{3}{c}{10\%} \\
            Model & Pr. & Rec. & F1 & Pr. & Rec. & F1 & Prec. & Rec. & F1 \\
            \midrule
            \midrule
            SpanBERT & 0.0 & 0.0 & 0.0 \small{$\pm 0.0$} & 36.3 & 23.9 & 28.8 \small{$\pm 13.5$} & 3.2 & 1.1 & 1.6 \small{$\pm 20.7$} \\
            RoBERTa  &  56.8 & 4.1 & 7.7 \small{$\pm 3.6$}  & 52.8 & 34.6 & 41.8 \small{$\pm 3.3$} & 61.0 & 50.3 & 55.1 \small{$\pm 0.8$} \\
            K-Adapter &  73.8 & 7.6 & 13.8 \small{$\pm 3.4$} & 56.4 & 37.6 & 45.1 \small{$\pm 0.1$} & 62.3 & 50.9 & 56.0 \small{$\pm 1.3$} \\
            LUKE & 61.5 & 9.9 & 17.0 \small{$\pm 5.9$} & 57.1 & 47.0 & 51.6 \small{$\pm 0.4$} & 60.6 & 60.6 & 60.6 \small{$\pm 0.4$} \\
            \midrule
            NLI\textsubscript{RoBERTa} (ours)& 56.6 & 55.6 & 56.1 \small{$\pm 0.0$} & 60.4 & 68.3 & 64.1 \small{$\pm 0.2$} & \textbf{65.8} & 69.9 & 67.8 \small{$\pm 0.2$} \\
            NLI\textsubscript{DeBERTa}  (ours) & \textbf{59.5} & \textbf{68.5} & \textbf{63.7} \small{$\pm 0.0$} & \textbf{64.1} & \textbf{74.8} & \textbf{69.0} \small{$\pm 0.2$} & 62.4 & \textbf{74.4} & \textbf{67.9} \small{$\pm 0.5$}\\
            \bottomrule
        \end{tabular}
    }
    \caption{Few-shot scenario results with 1\%, 5\% and 10\% of training data. Precision, Recall and F1 score (standard deviation) of the median of 3 different runs are reported. Top four rows for third-party RE systems run by us. }
    \label{tab:few_shot}
\end{table*}

\section{Results} \label{sec:results}


\subsection{Zero-Shot}

\begin{figure}
    \centering
    \resizebox{0.5\textwidth}{!}{
        \includegraphics{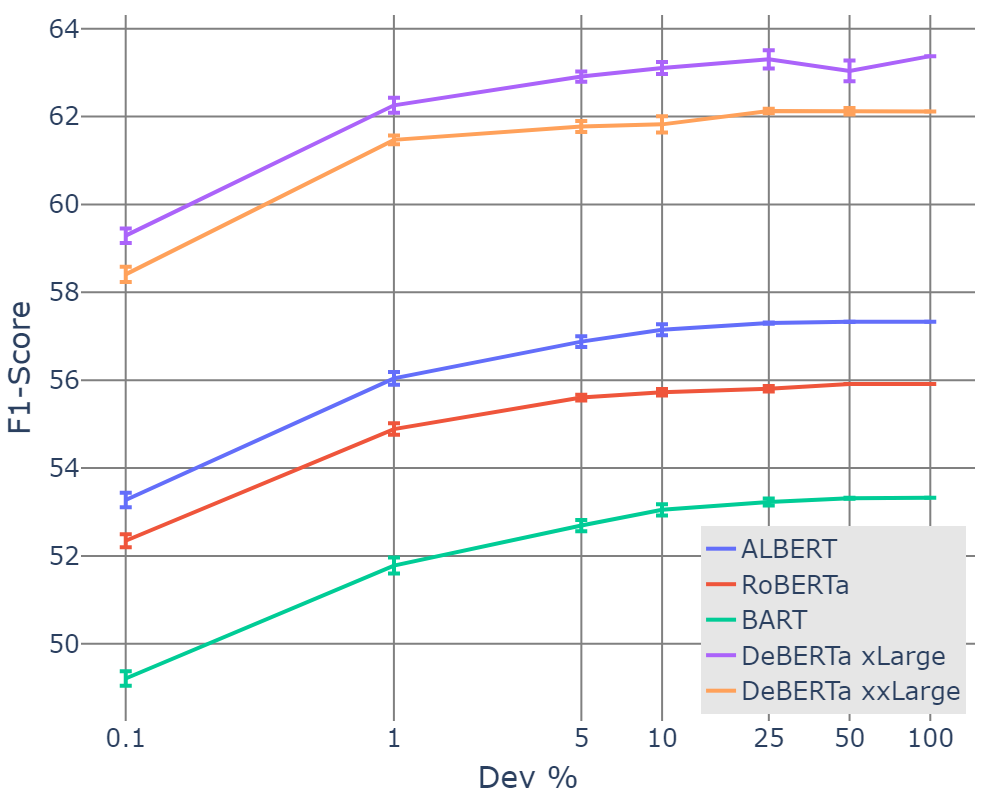}
    }
    \caption{Zero-shot scenario results. Mean F1 and standard error scores when setting $\mathcal{T}$ on increasing number of development examples. }
    \label{fig:threshold_estimation}
\end{figure}


Table \ref{tab:zero_shot} shows the results for different pre-trained NLI models, as well as the number of parameters and the MNLI \textit{matched} accuracy. 
These results were obtained by using the threshold for negative relations, as we found that it works substantially better than the no-relation template alternative (cf. Section \ref{ssed:zeroshot}). 
For instance, RoBERTa yields an F1 of 30.1\footnote{Results ommitted from Table \ref{tab:zero_shot} for brevity.} well below the 45.7 when using the default threshold ($\mathcal{T}=0.5$). Overall we see an excellent zero-shot performance across all the models and settings proving that the approach is robust and model agnostic.

Regarding \textbf{pre-trained models}, the best F1 scores are obtained by the two DeBERTa v2 models, which also score the best on the MNLI dataset. Note that all the models achieve similar scores on MNLI, but small differences in MNLI  result in large performance gaps when they come to RE, e.g. the 1.5 difference in MNLI between RoBERTa and DeBERTa becomes 7 points in No Dev. and 1\% Dev. We think the larger differences in RE are due to the generalization ability of some of the larger models to domain and task differences.

The table includes the results for different values of the $\mathcal{T}$ hyperparameter. In the most challenging setting, with default $\mathcal{T}$, the results are worst, with at most 57.8 F1. However, using as few as 2 examples per relation in average (1\% Dev. setting) the results improve significantly.

We  performed further experiments using larger amounts of development data to tune $\mathcal{T}$. Figure \ref{fig:threshold_estimation} shows that, for all models, the most significant improvement occurs at the interval [0\%, 1\%) and that the interval [1\%, 100\%] is almost flat. The best results with all development data is 63.4\%, only 0.6 points better than using 1\% of development. These results show clearly that a small number of examples suffice to set an optimal threshold. 


\subsection{Few-Shot}



Table \ref{tab:few_shot} shows the results of competing RE systems and our systems on the few-shot scenario. We report the median and standard deviation across 3 different runs. 
The competing RE methods suffer a large performance drop, specially for the smallest training setting. 
For instance, the SpanBERT system \cite{joshi-etal-2020-spanbert} has difficulties to converge, even with the 10\% of data setting. Both K-Adapter \cite{wang2020kadapter} and LUKE \cite{yamada-etal-2020-luke} improve over the RoBERTa system \cite{wang2020kadapter} in all three settings, but they are well below our NLI\textsubscript{RoBERTa} system, with improvements of 48, 22 and 13 points against the baseline in each setting. We also report our method based on DeBERTa\textsubscript{xLarge}, which is specially effective in the smaller settings. 

We would like to note that the zero-shot NLI\textsubscript{RoBERTa} system (1\% Dev) is comparable in terms of F1 score to a vanilla RoBERTa trained with 10\% of the training data. That is, 54 templates (10.5 hours, plus 23 development examples are roughly equivalent to 6800 annotated examples\footnote{Unfortunately we could not find the time estimates for annotating examples.} for training (plus 2265 development) .

\subsection{Full training}

\begin{table}
    \centering
    \resizebox{0.4\textwidth}{!}{
        \begin{tabular}{l|ccc}
            \toprule
            Model & Pr. & Rec. & F1 \\
            \midrule
            \midrule
            SpanBERT & 70.8 & 70.9 & 70.8 \\
            RoBERTa  & 70.2 & 72.4 & 71.3 \\
            K-Adapter & 70.1 & 74.0 & 72.0 \\
            LUKE & 70.4 & 75.1 & 72.7 \\
            \midrule
            NLI\textsubscript{RoBERTa} (ours) & 71.6 & 70.4 & 71.0 \\
            NLI\textsubscript{DeBERTa} (ours) & \textbf{72.5} & \textbf{75.3} & \textbf{73.9} \\
            \bottomrule
        \end{tabular}
    }
    \caption{Full training results (TACRED). Top four rows for third-party RE systems as reported by authors.}
    \label{tab:full_ft}
\end{table}

Some zero-shot and few-shot systems are not able to improve results when larger amounts of training data are available. 
Table \ref{tab:full_ft} reports the results when the whole train and development datasets are used, which is comparable to official results on TACRED. Focusing on our NLI\textsubscript{RoBERTa} system, and comparing it to the results in Table \ref{tab:few_shot}, 
we can see that it is able to effectively use the additional training data, improving from 67.9 to 71.0. When compared to a traditional RE system, it performs on a par to RoBERTa, and a little behind K-Adapter and LUKE, probably due to the infused knowledge which our model is not using. These results show that our model keeps improving with additional data and that it is competitive when larger amounts of training is available. The results of NLI\textsubscript{DeBERTa} show that our model can benefit from larger and more effective pre-trained NLI systems even in full training scenarios, and in fact achieves the best results to date on the TACRED dataset. 



\subsection{Data augmentation results}

\begin{table}
    \centering
    \resizebox{0.45\textwidth}{!}{
        \begin{tabular}{l|cccc} 
            \toprule
            Model & 0\% & 1\% & 5\% & 10\% \\
            \midrule
            \midrule
            RoBERTa & - & 7.7 & 41.8 & 55.1 \\
            \midrule
            + Zero-Shot DA & \textbf{56.3} & \textbf{58.4} & 58.8 & 59.7 \\ 
            + Few-Shot DA & - & \textbf{58.4} & \textbf{64.9} & \textbf{67.7}\\ 
            \bottomrule
        \end{tabular}
    }
    \caption{Data Augmentation scenario results (F1) for different gold training sizes. Silver annotations by the zero-shot and few-shot NLI\textsubscript{RoBERTa} model.}
    \label{tab:data_aug}
\end{table}

In this section we explore whether our NLI-based system can produce high-quality silver data which can be added to a small amount of gold data when training a traditional supervised RE system, e.g. the RoBERTa baseline \cite{wang2020kadapter}. 
Table \ref{tab:data_aug} reports the F1 results on the data augmentation scenario for different amounts of gold training data. 
Overall, we can see that both our zero-shot and few-shot methods\footnote{The zero-shot 1\% Dev model is used in all data augmentation experiments, while the few-shot method changes to use the available data at each run (1\%, 5\% and 10\%), both with RoBERTa} provide good quality silver data, as they improve significantly over the baseline in all settings. 
Although the zero-shot and few-shot methods yield the same result with 1\% of training data, the few-shot model is better in the rest of training regimes, showing that it can effectively use the available training data in each case to provide better quality silver data. If we compare the results in this table with those of the respective NLI-based system with the same amount of gold training instances (Tables \ref{tab:zero_shot} and \ref{tab:few_shot}) we can see that the results are comparable, showing that our NLI-based system and a traditional RE system trained with silver annotations have comparable performance. A practical advantage of a traditional RE system trained with our silver data is that is easier to integrate on available pipelines, as one just needs to download the trained Transformer model. It also makes it easy to check additive improvements in the RE method.



\section{Analysis}

\begin{table}
    \centering
    \resizebox{0.5\textwidth}{!}{
        \begin{tabular}{lll|cc}
            \toprule
            Model & \multicolumn{2}{c|}{Scenario} &  P & PvsN \\
            \midrule
            \midrule
            \multirow{4}{*}{NLI\textsubscript{DeBERTa}} & \multirow{2}{*}{Zero-Shot} & No Dev & 85.6 & 59.5  \\
            & & 1\% Dev & 85.6 & 67.7 \\
            & Few-Shot & 5\% & 89.7 & 74.5 \\ 
            & Full train & - & 92.2 & 77.8 \\ 
            \midrule
            \multirow{2}{*}{LUKE} & Few-Shot & 5\%  & 69.3 & 63.4\\ 
              & Full train & -  & 90.2 & 77.3 \\ 
            \bottomrule
        \end{tabular}
    }
    \caption{Performance of selected systems and scenarios on two metrics: the binary task of detecting a positive relation vs. no-relation (PvsN column, F1) and detecting the correct relation among positive cases (P, F1).}
    \label{tab:PvsNP}
\end{table}

Relation extraction can be analysed according to two auxiliary metrics: the binary task of detecting a positive relation vs. no-relation, and the multi-class problem of detecting which relation holds among positive cases (that is, discarding no-relation instances from test data). Table \ref{tab:PvsNP} shows the results of a selection of systems and scenarios. The first rows compare the performance of our best system, NLI\textsubscript{DeBERTa}, across four scenarios, while the last two rows show the results for LUKE in two scenarios. The zero-shot No dev. system is very effective when discriminating the relation among positive examples (P column), only 7 points below the fully trained system, while it lags well behind when discriminating positive vs. negative, 18 points. The use of a small development data for tuning the $\mathcal{T}$ threshold closes the gap in PvsN, as expected, but the difference is still 10 points. All in all, these numbers show that our zero-shot system is very effective discriminating among positive examples, but that it still lags behind when detecting no-relation cases. Overall, the figures show the effectiveness of our methods in low data scenarios on both metrics.


\begin{figure*}[ht]
    \centering
    \resizebox{.77\textwidth}{!}{
        \includegraphics{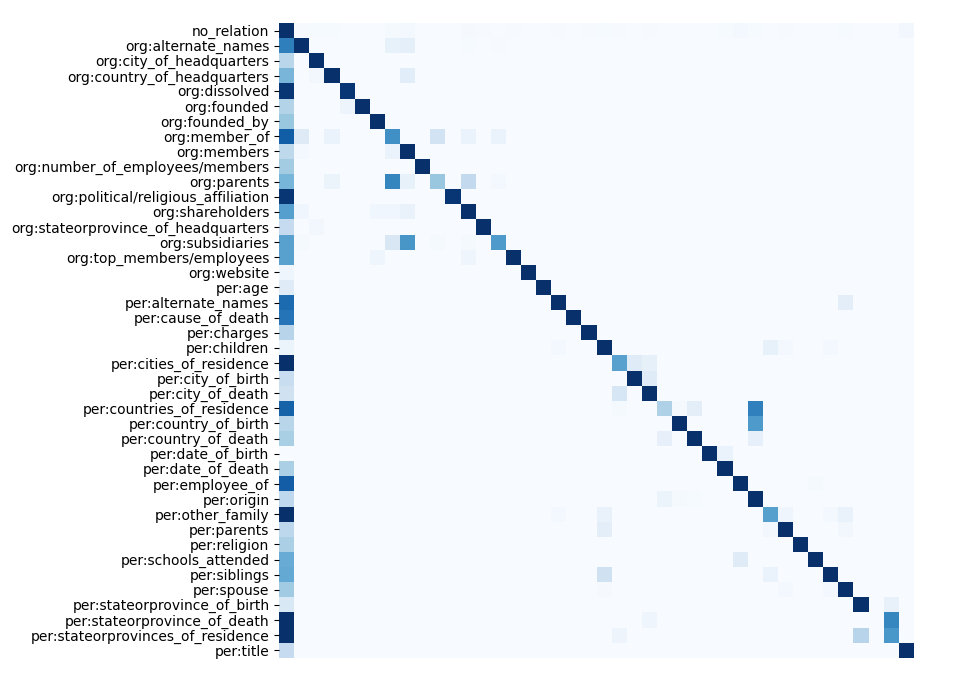}
    }
    \caption{Confusion matrix of our NLI\textsubscript{DeBERTa} zero-shot system on the development dataset. The rows represent the true labels and the columns the predictions. The matrix is rowise normalized (recall in the diagonal).}
    \label{fig:cm}
\end{figure*}

\paragraph{Confusion analysis} In supervised models some classes (relations) are better represented in training than others, usually due to data imbalance. Our system instead, represents each relations as a set of templates, which at least on a \textbf{zero-shot} scenario, should not be affected by data imbalance. The strong diagonal in the confusion matrix (Fig. \ref{fig:cm}) shows that our the model is able to discriminate properly between most of the relations (after all it achieves 85.6\% accuracy, cf. Table \ref{tab:PvsNP}), with exception of the no-relation column, which was expected. Regarding the confusion between actual relations, most of them are about \textbf{overlapping relations}, as expected. For instance, \textsc{org:member\_of} and \textsc{org:parents} both involve some organization A being part or member of some other organization B, where \textsc{org:members} is different from \textsc{org:parents}  in that correct fillers are distinct entities that are generally capable of autonomously ending their membership with the assigned organization\footnote{Description extracted from the guidelines.}. Something similar occurs between \textsc{org:members} and \textsc{org:subsidiaries}. Another reason for confusion happens when \textbf{two or more relations exist concurrently}, as in \textsc{per:origin}, \textsc{per:country\_of\_birth} and \textsc{per:country\_of\_residence}. Finally, the model scores low on \textsc{per:other\_family}, which is a bucket of many specific relations where only a handful were actually covered by the templates.

\section{Conclusions}
In this work we reformulate relation extraction as an entailment problem, and explore to what extent simple hand-made verbalizations are effective.  The creation of templates is limited to 15 minutes per relation, and yet allows for excellent results in zero- and few-shot scenarios. Our method makes effective use of available labeled examples, and together with larger LMs produces the best results on TACRED to date. 
Our analysis indicates that the main performance difference against supervised models comes from discriminating no-relation examples, as the performance among positive examples equals that of the best supervised system using the full training data. We also show that our method can be used effectively as a data-augmentation method to provide additional labeled examples. For the future we would like to investigate better methods for detecting no-relation in zero-shot settings.

\section*{Acknowledgements}
Oscar is funded by a PhD grant from the Basque Government (PRE\_2020\_1\_0246). This work is based upon work partially supported via the IARPA  BETTER Program contract No. 2019-19051600006 (ODNI, IARPA), and by the Basque Government (IXA excellence research group IT1343-19). 

\bibliography{anthology,custom}
\bibliographystyle{acl_natbib}


  \clearpage

\appendix

\setcounter{table}{0}

\section{Pre-Trained models}
\label{sec:pretrained}

The pre-trained NLI models we have tested from the Transformers library are the next:
\begin{itemize}
    \item ALBERT: \textit{ynie/albert-xxlarge-v2-snli\_mnli\ \_fever\_anli\_R1\_R2\_R3-nli}
    \item RoBERTa: \textit{roberta-large-mnli}
    \item BART: \textit{facebook/bart-large-mnli}
    \item DeBERTa v2 xLarge: \textit{microsoft/deberta-v2-xlarge-mnli}
    \item DeBERTa v2 xxLarge: \textit{microsoft/deberta-v2-xxlarge-mnli}
\end{itemize}

\section{Experimental details}
\label{sec:experimental_details}

We carried out all the experiments on a single Titan V (16GB) except for the fine-tuning of DeBERTa, that has been done on a cluster of 4 Titan V100 (32GB). The average inference time for the zero and few-shot experiments is between 1h and 1.5h. The time needed for fine-tuning the NLI systems was at most 2.5h for RoBERTa and 5h for DeBERTa. All the experiments were done with mixed precision to speed up the overall runtime.

The whole hyperparameter settings used for fine-tuning NLI\textsubscript{RoBERTa} and NLI\textsubscript{DeBERTa} are listed below:

\begin{itemize}
    \item \textbf{Train epochs:} 2
    \item \textbf{Warmup steps:} 1000
    \item \textbf{Learning-rate:} 4e-6
    \item \textbf{Batch-size:} 32
    \item \textbf{FP16 training}
    \item \textbf{Seeds:} \{0, 24, 42\}
    
\end{itemize}

Note that we are fine-tuning an already trained NLI system so we kept the number of epochs and learning-rate low. The rest of state-of-the-art systems were trained using the hyperparameters reported by the authors.

\section{TACRED templates}
\label{sec:templates}

This section describes the templates used in the TACRED experiments. We performed all the experiments using the templates showed in Tables \ref{tab:per_templates} (for PERSON relations) and \ref{tab:org_templates} (for ORGANIZATION relations). These templates were manually created based on the TAC KBP Slot Descriptions\footnote{\url{https://tac.nist.gov/2014/KBP/ColdStart/guidelines/TAC_KBP_2014_Slot_Descriptions_V1.4.pdf}} (annotation guidelines). Besides the templates, we also report the valid argument types that are accepted on each relation.

\begin{table*}[t]
    \centering
    \resizebox{\textwidth}{!}{
        \begin{tabular}{rcl}
            \toprule
            Relation & Templates & Valid argument types \\
            \midrule
            per:alternate\_names & \{subj\} is also known as \{obj\} & PERSON, MISC \\
            per:date\_of\_birth & \{subj\}'s birthday is on \{obj\} & DATE \\
             & \{subj\} was born on \{obj\} & \\
            per:age & \{subj\} is \{obj\} years old & NUMBER, DURATION \\
            per:country\_of\_birth & \{subj\} was born in \{obj\} & COUNTRY \\
            per:stateorprovince\_of\_birth & \{subj\} was born in \{obj\} & STATE\_OR\_PROVINCE \\
            per:city\_of\_birth & \{subj\} was born in \{obj\} & CITY, LOCATION \\
            per:origin & \{obj\} is the nationality of \{subj\} & NATIONALITY, COUNTRY, LOCATION \\
            per:date\_of\_death & \{subj\} died in \{obj\} & DATE \\
            per:country\_of\_death & \{subj\} died in \{obj\} & COUNTRY \\
            per:stateorprovince\_of\_death & \{subj\} died in \{obj\} & STATE\_OR\_PROVINCE \\
            per:city\_of\_death & \{subj\} died in \{obj\} & CITY, LOCATION \\
            per:cause\_of\_death & \{obj\} is the cause of \{subj\}'s death & CAUSE\_OF\_DEATH \\
            per:countries\_of\_residence & \{subj\} lives in \{obj\} & COUNTRY, NATIONALITY \\
             & \{subj\} has a legal order to stay in \{obj\} & \\
            per:statesorprovinces\_of\_residence & \{subj\} lives in \{obj\} & STATE\_OR\_PROVINCE \\
             & \{subj\} has a legal order to stay in \{obj\} & \\
            per:city\_of\_residence & \{subj\} lives in \{obj\} & CITY, LOCATION \\
             & \{subj\} has a legal order to stay in \{obj\} & \\
            per:schools\_attended & \{subj\} studied in \{obj\} & ORGANIZATION \\
             & \{subj\} graduated from \{obj\} & \\
            per:title & \{subj\} is a \{obj\} & TITLE \\
            per:employee\_of & \{subj\} is a member of \{obj\} & ORGANIZATION \\
            per:religion & \{subj\} belongs to \{obj\} & RELIGION \\
             & \{obj\} is the religion of \{subj\} & \\
             & \{subj\} believe in \{obj\} & \\
            per:spouse & \{subj\} is the spouse of \{obj\} & PERSON \\
             & \{subj\} is the wife of \{obj\} & \\
             & \{subj\} is the husband of \{obj\} & \\
            per:children & \{subj\} is the parent of \{obj\} & PERSON \\
             & \{subj\} is the mother of \{obj\} & \\
             & \{subj\} is the father of \{obj\} & \\
             & \{obj\} is the son of \{subj\} & \\
             & \{obj\} is the daughter of \{subj\} & \\
            per:parents & \{obj\} is the parent of \{subj\} & PERSON \\
             & \{obj\} is the mother of \{subj\} & \\
             & \{obj\} is the father of \{subj\} & \\
             & \{subj\} is the son of \{obj\} & \\
             & \{subj\} is the daughter of \{obj\} & \\
            per:siblings & \{subj\} and \{obj\} are siblings & PERSON \\
             & \{subj\} is brother of \{obj\} & \\
             & \{subj\} is sister of \{obj\} & \\
            per:other\_family & \{subj\} and \{obj\} are family & PERSON \\
             & \{subj\} is a brother in law of \{obj\} & \\
             & \{subj\} is a sister in law of \{obj\} & \\
             & \{subj\} is the cousin of \{obj\} & \\
             & \{subj\} is the uncle of \{obj\} & \\
             & \{subj\} is the aunt of \{obj\} & \\
             & \{subj\} is the grandparent of \{obj\} & \\
             & \{subj\} is the grandmother of \{obj\} & \\
             & \{subj\} is the grandson of \{obj\} & \\
             & \{subj\} is the granddaughter of \{obj\} & \\
            per:charges & \{subj\} was convicted of \{obj\} & CRIMINAL\_CHARGE \\
             & \{obj\} are the charges of \{subj\} & \\

            \bottomrule
        \end{tabular}
    }
    \caption{Templates and valid arguments for PERSON relations.}
    \label{tab:per_templates}
\end{table*}

\begin{table*}
    \centering
    \resizebox{\textwidth}{!}{
        \begin{tabular}{rcl}
            \toprule
            Relation & Templates & Valid argument types \\
            \midrule
            org:alternate\_names & \{subj\} is also known as \{obj\} & ORGANIZATION, MISC \\
            org:political/religious\_affiliation & \{subj\} has political affiliation with \{obj\} & RELIGION, IDEOLOGY \\
             & \{subj\} has religious affiliation with \{obj\} & \\
            org:top\_memberts/employees & \{obj\} is a high level member of \{subj\} & PERSON \\
             & \{obj\} is chairman of \{subj\} & \\
             & \{obj\} is president of \{subj\} & \\
             & \{obj\} is director of \{subj\} & \\
            org:number\_of\_employees/members & \{subj\} employs nearly \{obj\} people & NUMBER \\
             & \{subj\} has about \{obj\} employees & \\
            org:members & \{obj\} is member of \{subj\} & ORGANIZATION, COUNTRY \\
             & \{obj\} joined \{subj\} & \\
            org:subsidiaries & \{obj\} is a subsidiary of \{subj\} & ORGANIZATION, LOCATION \\
             & \{obj\} is a branch of \{subj\} & \\
            org:parents & \{subj\} is a subsidiary of \{obj\} & ORGANIZATION, COUNTRY \\
             & \{subj\} is a branch of \{obj\} & \\
            org:founded\_by & \{subj\} was founded by \{obj\} & PERSON \\
             & \{obj\} founded \{subj\} & \\
            org:founded & \{subj\} was founded in \{obj\} & DATE \\
             & \{subj\} was formed in \{obj\} & \\
            org:dissolved & \{subj\} existed until \{obj\} & DATE \\
             & \{subj\} disbanded in \{obj\} & \\
             & \{subj\} dissolved in \{obj\} & \\
            org:country\_of\_headquarters & \{subj\} has its headquarters in \{obj\} & COUNTRY \\
             & \{subj\} is located in \{obj\} & \\
            org:stateorprovince\_of\_headquarters & \{subj\} has its headquarters in \{obj\} & STATE\_OR\_PROVINCE \\
             & \{subj\} is located in \{obj\} & \\
             org:city\_of\_headquarters & \{subj\} has its headquarters in \{obj\} & CITY, LOCATION \\
             & \{subj\} is located in \{obj\} & \\
            org:shareholders & \{obj\} holds shares in \{subj\} & ORGANIZATION, PERSON \\
            org:website & \{obj\} is the URL of \{subj\} & URL \\
             & \{obj\} is the website of \{subj\} & \\
            \bottomrule
        \end{tabular}
    }
    \caption{Templates and valid arguments for ORGANIZATION relations.}
    \label{tab:org_templates}
\end{table*}

\end{document}